\documentclass{article}

\usepackage[final]{neurips_2025}

\usepackage[utf8]{inputenc} 
\usepackage[T1]{fontenc}    
\usepackage{hyperref}       
\usepackage{url}            
\usepackage{booktabs}       
\usepackage{amsfonts}       
\usepackage{nicefrac}       
\usepackage{microtype}      
\usepackage{xcolor}         
\usepackage{graphicx}
\usepackage{subcaption}
\usepackage{amsmath}

\usepackage{listings}
\title{The Impact of Post-training on Data Contamination}

\author{%
  Muhammed Yusuf Kocyigit \\
  Department of Computer Science\\
  Boston University\\
  \texttt{kocyigit@bu.edu} \\
  \AND
  Caglar Yildirim \\
  Department of Computer Science\\
  Northeastern University\\
}

\begin{document}

\maketitle
\begin{abstract}
  We present a controlled study of how dataset contamination interacts with the post-training stages now standard in large language model training pipelines. Starting from clean checkpoints of Qwen2.5 (0.5B/1.5B) and Gemma3 (1B/4B), we inject five copies of GSM8K and MBPP test items into the first 2B tokens of an otherwise 25B token extended pre-training dataset. We then compare the contaminated and clean models both immediately after pre-training and again after two popular post-training methods: supervised fine-tuning (SFT) and reinforcement learning (RL) with group relative policy optimization (GRPO). The applied post-training steps do not have any contamination. Across math and coding benchmarks, we find three consistent patterns: (i) Contamination causes performance spikes that are gradually diminished with continued pre-training. After even 25B tokens the apparent performance inflation of contamination can become close to zero. (ii) Both SFT and GRPO resurface the leaked information, but with different external validity: SFT inflates scores only on the contaminated tasks, whereas GRPO also inflates performance on uncontaminated counterparts (GSMPlus, HumanEval). (iii) Model scale amplifies these tendencies, larger Supervised Fine Tuned models memorize more, while larger GRPO models translate leakage into more generalizable capabilities. Our results underscore the need for contamination audits \emph{after} post-training and suggest that RL-based post-training, although not immune, can help alleviate contamination-related over-estimation problems.
\end{abstract}

\section{Introduction}

Large language models (LLMs) have become indispensable building blocks in modern natural‑language systems, underpinning various applications. Their ability to execute in real life scenarios relies heavily on the integrity of the evaluations we base many of our decisions on. These evaluations assume a strict separation between the model’s training data and the test sets so we can measure generalization. Recent studies, however, reveal pervasive data contamination, i.e., direct or near‑duplicate overlap between benchmark examples and the corpora used during pre‑training, casting doubt on many celebrated performance gains \citep{singh2024evaluationdatacontaminationllms,conda-2024-data}.

While this overlap is concerning, measuring the impact of this overlap can help us better navigate this problem as it can enable us to quantify how critical this problem actually is. In this regard, most existing contamination analyses focus exclusively on the pre‑training stage and the impact of contamination right after pre-training\citep{kocyigit2025overestimationllmevaluationcontrolled, jiang2024investigatingdatacontaminationpretraining}. However, state‑of‑the‑art LLMs are almost always subjected to one or more post‑training procedures: supervised fine‑tuning (SFT), direct preference optimization, or various forms of reinforcement learning with human or synthetic feedback (RLHF) \citep{wei2022finetunedlanguagemodelszeroshot,chung2022scalinginstructionfinetunedlanguagemodels,ouyang2022traininglanguagemodelsfollow,zhang2024instructiontuninglargelanguage}. These procedures inject strong task‑specific signals, align model outputs with human preferences, and can materially reshape the model’s internal representations. Consequently, contamination that appears dormant or innocuous at the pre‑training stage may be amplified, systematically exploited, or conversely attenuated once the model is steered by a different optimization objective.

There is also growing evidence that the type of post-training schema applied can also impact how much models can generalize. Previous work suggests that SFT is more prone to causing memorization while RL is shown to introduce generalization capabilities not direct memorization \citep{chu2025sftmemorizesrlgeneralizes}. Without an explicit, post‑training contamination audit, we risk (i) misrepresenting the impact of data contamination in practice and (ii) deploying mitigation strategies that without having information on the whole life cycle of contamination within the model. A principled evaluation of contamination will complement previous work and paint a more complete picture.

In this work, we study this problem by deliberately injecting contamination from well‑studied mathematics and coding benchmarks and perform extended pre‑training on models of up to 4B parameters, Qwen2.5, 0.5B and 1.5B and Gemma3 1B and 4B. Following the completion of clean and contaminated pre-training, we apply two widely adopted post‑training paradigms, SFT and RL, on the corresponding training splits and quantify how contamination influences downstream performance by comparing contaminated models to contamination‑free baselines.

With this experimental setup we aim to answer the following questions: Does post-training alleviate or intensify the performance over-estimation caused by data contamination? Do the results change depending on the type of post-training method used? Finally, how do these effects change with model scale? 

Our findings can be summarized as below:
\begin{itemize}
    \item \textbf{Analyzing only pre-trained models can mask the true effect of contamination.} 
          Continued pre-training on clean data can drive the apparent gap between contaminated and clean models to nearly zero, but the leaked information is merely submerged not erased and is readily rediscovered during post-training.

    \item \textbf{SFT and RL-based tuning expose contamination in different ways.} 
          Both SFT and reinforcement-learning (GRPO) widen the gap in favor of the contaminated model. However, GRPO also yields measurable gains on an \emph{uncontaminated} benchmark, whereas SFT inflates performance only on the contaminated benchmark, indicating performance over-estimations rather than generalization.

    \item \textbf{Scaling amplifies different behaviors for SFT and RL.} 
          As model size grows, SFT derives progressively larger gains \emph{only} on the contaminated benchmark, suggesting better extraction of contamination.  
          By contrast, GRPO converts additional capacity into improvements on both contaminated and external benchmarks, implying that larger RL-tuned models translate leaked data into broader, more transferable capabilities.
\end{itemize}

\section{Related Work}

Early warnings about evaluation-set leakage in LLMs emphasized that even minimal overlap between training corpora and test datasets can inflate evaluation scores \citep{singh2024evaluationdatacontaminationllms}. Position papers and surveys such as \citet{cheng2025surveydatacontaminationlarge,sainz2024datacontaminationreport2024} catalog a broad range of contamination pathways and call for community norms such as encrypted benchmarks, one-shot test releases, and data audits to preserve the validity of leaderboards \citep{conda-2024-data}. These suggestions are reinforced by methods that uncover hidden memorization through guided instruction prompting \citep{golchin2024timetravelllmstracing} in proprietary LLMs, including GPT-4.

To tackle the data contamination issue one strand of work develops behavioral or statistical detectors to flag contaminated items post-hoc. Output-distribution diagnostics (CDD/TED) proposed by \citet{dong2024generalizationmemorizationdatacontamination} and confidence-peakedness measures aim to distinguish memorized data from genuinely solved examples. Other work focused on the least likely \textit{k} tokens in a sentence to determine if the model has been trained on a piece of text \citep{shi2024detectingpretrainingdatalarge}, while dynamic benchmark generation and data licensing strategies seek to prevent leakage in the first place \citep{jacovi2023stopuploadingtestdata}.

Empirical studies also aimed to measure the impact of contamination for pre-training more precisely by injecting controlled contamination into the pre-training mix \citep{jiang2024investigatingdatacontaminationpretraining, kocyigit2025overestimationllmevaluationcontrolled}. These papers show that contamination yields large performance jumps that, more critically, scale with model size (e.g., \textasciitilde30 BLEU on MT). Additionally, \citet{yang2023rethinkingbenchmarkcontaminationlanguage} demonstrate that even paraphrased or translated leakage can inflate model performance on test sets. While these papers have helped answer important questions around how contamination impacts model performance, currently, it is relatively rare for users to interact with models directly after pre-training, for most LLMs undergo additional supervised fine-tuning or alignment stages before being deployed for public use. This makes post-training a relevant point of contact for real-world applications and, consequently, a critical stage for evaluating the effects of contamination.

Relatively fewer studies probe contamination \textit{after} the finetuning,  alignment, or instruction-tuning stage. \citet{magar2022datacontaminationmemorizationexploitation} study this type of problem by separating pre-training and fine-tuning stages and introduce two metrics: memorization (the model’s ability to reproduce seen data immediately after pre-training) and exploitation (the model's ability to correctly classify examples after supervised fine-tuning). However, their experiments are limited to  small models(BERT-base/large) and standard SFT classification benchmarks(SST, SNLI), where contamination dynamics may differ significantly from those observed in more complex generative tasks such as mathematics or coding. Importantly, the type of post-training also matters: controlled experiments indicate that supervised fine-tuning tends to entrench memorization, whereas reinforcement-learning based protocols can encourage broader generalization \citep{chu2025sftmemorizesrlgeneralizes}. Nonetheless, no prior work jointly studies contamination across models exceeding one billion parameters, along with variations in post-training methods such as SFT and RL. Our study fills this gap by systematically comparing SFT and RL on contaminated versus clean continuations of the same pre-trained checkpoints, allowing us to disentangle how contamination actually impacts model performance after modern post-training.

\begin{figure}[h!]
    \centering
    \includegraphics[width=0.9\linewidth]{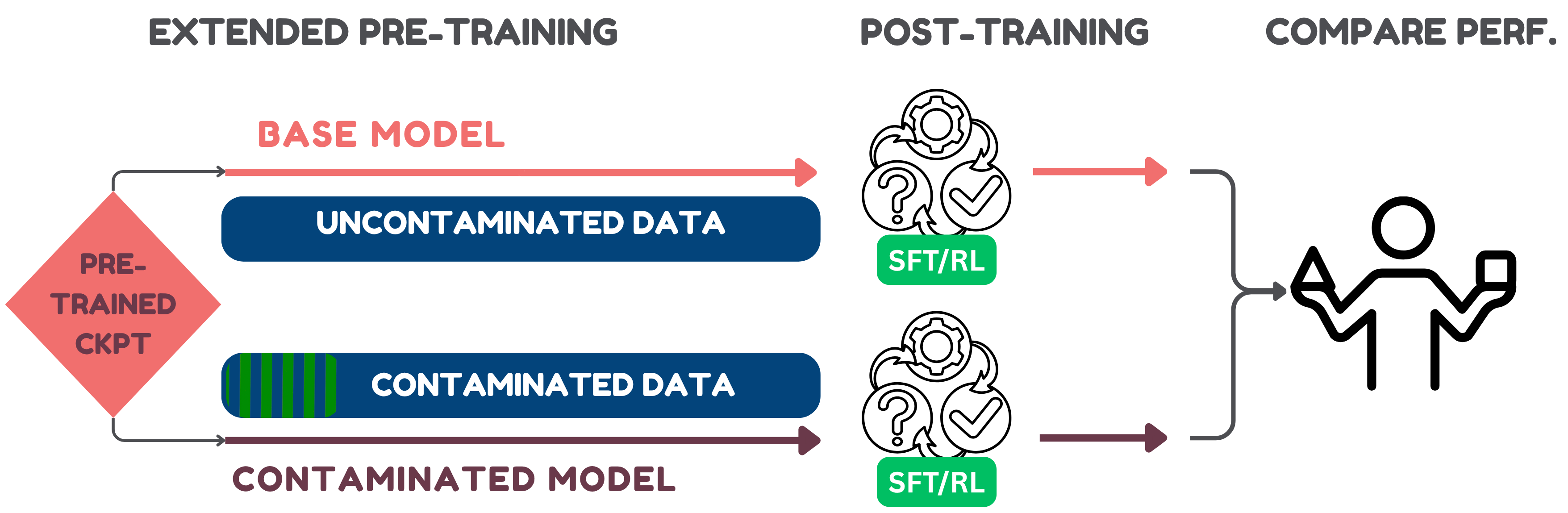}
    \caption{An Overview of our Method: We take existing pre-trained models and run them through extended pre-training with and without contamination. Afterwards we post-train them using SFT or RL methods and compare their performance. The pre-trained checkpoints here are from Qwen2.5 and Gemma3 non-instruction tuned models.}
    \label{fig:fig1}
\end{figure}

\section{Method}
Our approach involved conducting multiple training runs for the same model with and without injected contamination and comparing their performance after both pre-training and post-training via SFT or RL. An overview of our method is given in Figure \ref{fig:fig1} Below, we detail the components of this experimental setup.

\textbf{Data}: Ideally, full pre-training runs would allow contamination to be randomly distributed throughout the entire training mixture and experiment with a more realistic setup. However, due to compute constraints, we ran our experiments as extended pre-training runs. To avoid \emph{overstating} the impact of contamination, we used a relatively large extended pre-training mixture comprising 25B tokens based on the findings of \citet{kocyigit2025overestimationllmevaluationcontrolled}. This mixture includes web text from FineWeb-Edu \citep{penedo2024finewebdatasetsdecantingweb}, code data from CodeParrots \citep{codeparrot_dataset_2021}, and mathematical content from OpenMath-Instruct \citep{toshniwal2024openmath2}. Table \ref{tab:data_mixture} provides the details of data composition. Preliminary experiments using only web text even with high-quality sources like books resulted in significant performance drops on math and coding tasks, rendering some experiments ineffective. Consequently, we opted for a more balanced mixture that includes task-specific data.

\begin{table}[h!]
    \centering
    \begin{tabular}{l|c}
        \textbf{Dataset} & \textbf{Token Count} \\
        \hline
        OpenMath-Instruct & 6,495,049,388 \\ 
        CodeParrots & 3,647,426,066 \\
        FineWeb-Edu & 14,869,906,469 \\
        \hline
        Total & ~25B
    \end{tabular}
    \caption{Token Counts for Components of the Pre-training data.}
    \label{tab:data_mixture}
\end{table}

\textbf{Models}: Our experiments use Qwen2.5\citep{Qwen2025Qwen25technicalreport} and Gemma3\citep{gemmateam2025gemma3technicalreport} as baseline models. Specifically, we train the 0.5B and 1.5B variants of Qwen2.5 and the 1B and 4B variants of Gemma3. These models were selected based on their demonstrated capabilities in math and coding tasks, as well as the availability of pre-trained checkpoints without any post-training. Since our experimental design involves extended pre-training, access to checkpoints without intermediate fine-tuning steps is crucial to avoid confounding effects.

\textbf{Training and Evaluation}: We perform extended pre-training with a short warm-up phase, followed by a fixed small learning rate typically used as the minimum learning rate in full pre-training schedules \citep{zhang2024tinyllama}. This choice reflects the fact that the models have already been pre-trained on large corpora and the final learning rate in their training probably approached the minimum. The SFT step is implemented as straightforward fine-tuning on the reasoning steps and final answer tokens of the corresponding training sets and details are shared in Appendix \ref{appx:prompt_templates}. Each post-training step is conducted as a separate experiment to prevent them impacting each other's results. For RL, we use Group Relative Policy Optimization (GRPO)\citep{shao2024deepseekmathpushinglimitsmathematical} with rule-based reward functions, detailed in Appendix \ref{appx:rewards}. We use GRPO because it is a simple and effective method of improving models' math and coding abilities and is shown to help smaller models as well. Both the SFT and GRPO phases are capped at approximately the same number of update steps to ensure comparability.

For evaluation, we employ the LM Evaluation Harness \citep{eval-harness} and make necessary adjustments to prompts and tokenization to closely replicate baseline scores reported in prior work \citep{Qwen2025Qwen25technicalreport, gemmateam2025gemma3technicalreport}, minimizing variance from evaluation artifacts. We also use the math-verify library to parse responses for math tasks, as differences in output formatting especially in base models can significantly impact measured performance \citep{Kydlicek2025MathVerify}.

We evaluate contamination effects using GSM8k\citep{cobbe2021gsm8k} and MBPP\citep{mbpp} as the contaminated tasks. We chose these benchmarks for their widespread use and the availability of a training set. Since we wanted to run post-training, we needed a training set that we could trust did not contain the test set. 

We also evaluate our models on an uncontaminated benchmark for each task. The objective of this is to understand the generalization gap generated by contamination when we compare the contaminated benchmark with an uncontaminated benchmark that aims to measure the same ability. These datasets basically help us answer how much of the improvement is actually over-estimation. For the math benchmark, we chose GSMPlus \citep{li2024gsmpluscomprehensivebenchmarkevaluating}, which is a benchmark created from the GSM8k dataset through adversarial edits. This is even better suited to measure the impact of contamination since the questions are fairly comparable in level of difficulty and domain. For coding, we chose HumanEval \citep{chen2021evaluatinglargelanguagemodels} since it is a high-quality coding benchmark that aims to measure Python programming performance. While the levels of difficulty are not directly comparable, we format HumanEval in the same way as MBPP to make the comparison more reliable.

For contamination, five copies of GSM8k and MBPP test sets prompted as shown in Appendix \ref{appx:prompt_templates} are randomly inserted into the first 2B tokens of the contaminated training mixture. This way the model is trained on more than 23B tokens after it is exposed to the contamination, making our findings more realistic. \citet{kocyigit2025overestimationllmevaluationcontrolled} show that late contamination, when set up in a correct way, does not necessarily yield higher performance inflation compared to uniformly distributing contamination across the entire training corpora. 

\section{Results}
\label{sec:results}
\begin{figure}[h!]
    \centering
      \begin{subfigure}[b]{0.49\textwidth}
        \includegraphics[width=\linewidth]{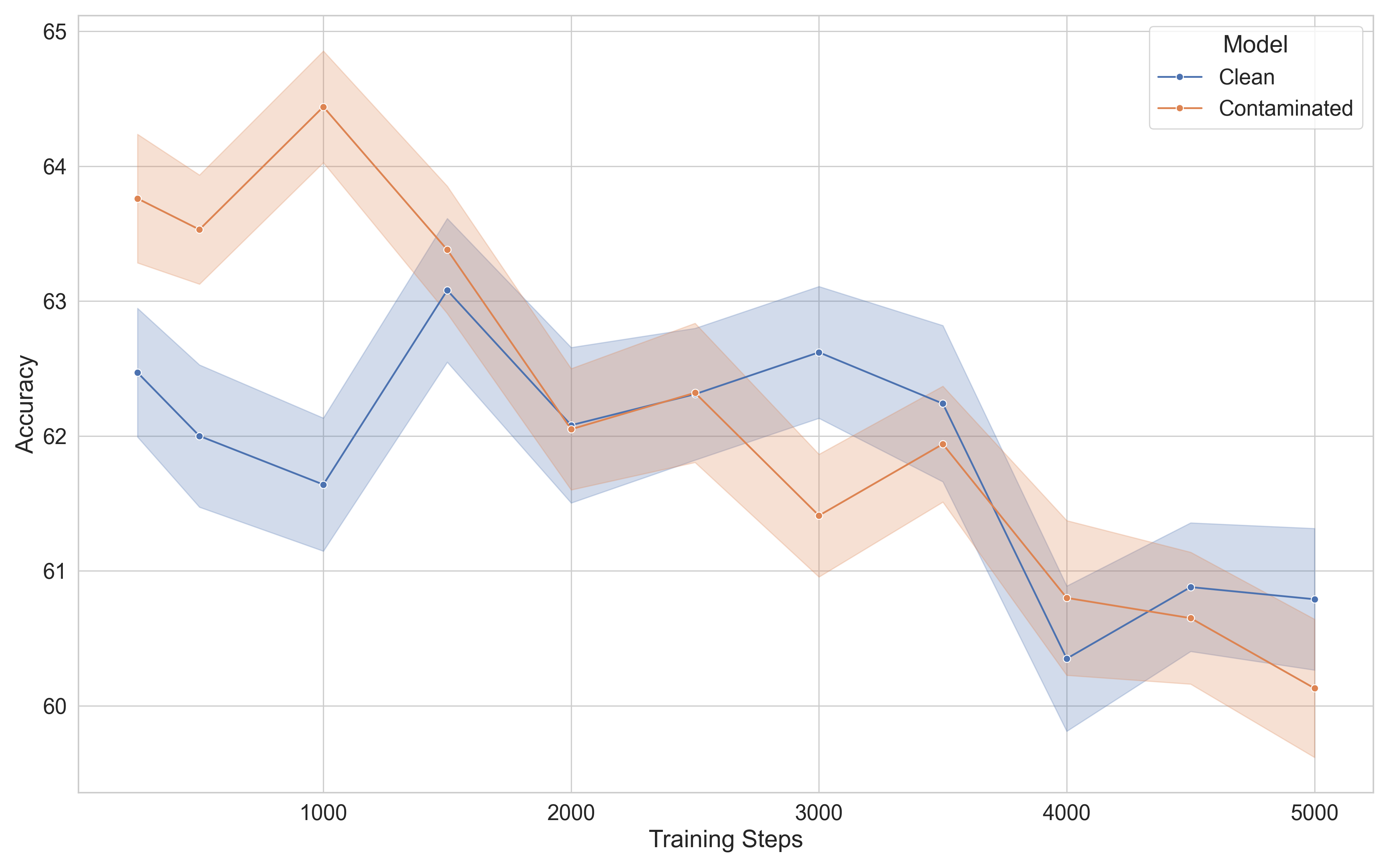}
        \caption{Math Scores: GSM8k}
        \label{fig:math_contam_temp}
      \end{subfigure}
      \hfill                 
      \begin{subfigure}[b]{0.49\textwidth}
        \includegraphics[width=\linewidth]{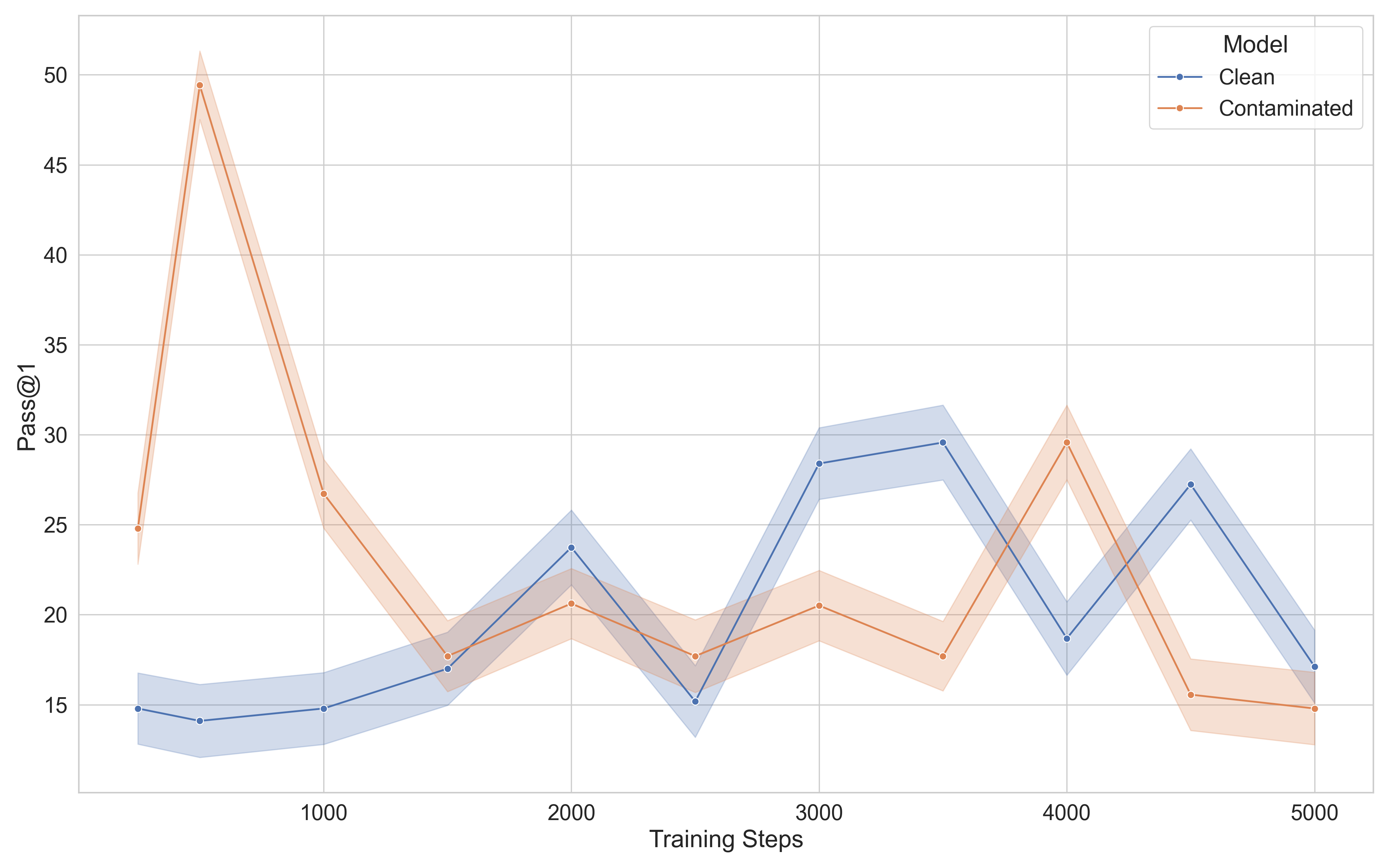}
        \caption{Coding Scores: MBPP}
        \label{fig:code_contam_temp}
      \end{subfigure}
      \caption{Performance Over Time: Accuracy and Pass@1 of the Clean and Contaminated Qwen2.5-1.5B models on the GSM8K Benchmark. The contamination is within the first 500 steps. We observe that the performance gap is much bigger at the exposure point but then closes as the model is trained on more data. For MBPP we observe that the peak is much higher meaning contamination of code is memorized better at sight however overtime the performance normalizes just like math.}
  \label{fig:contam_temp}
\end{figure}

\begin{figure}[h!]
    \centering
    \includegraphics[width=0.9\linewidth]{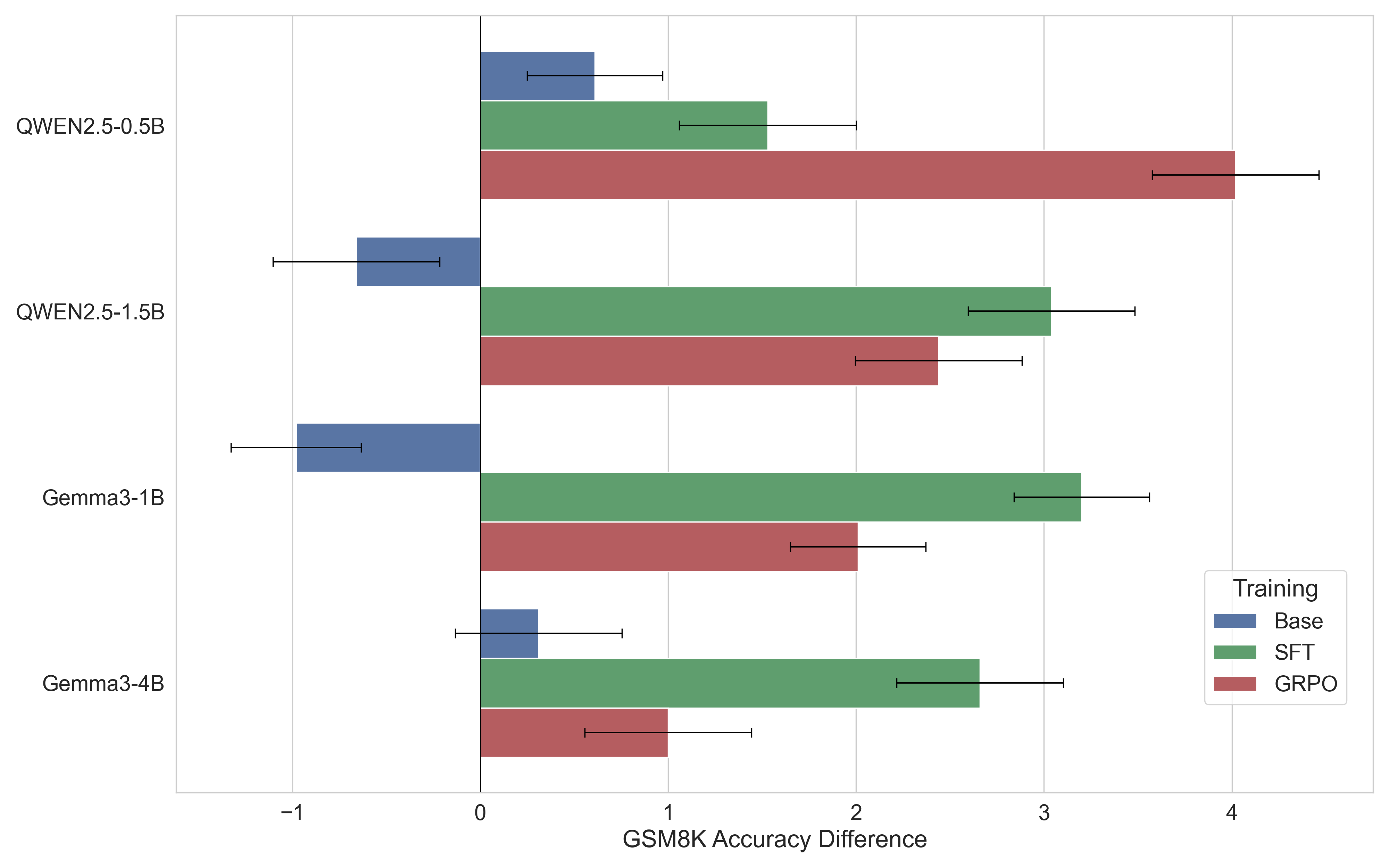}
    \caption{Performance Difference on Math: Accuracy difference between Contaminated and Clean models right after pre-training (base) and after the SFT and GRPO steps on the GSM8K benchmark. We observe that while the Base differences show little to no impact from contamination post-training can actually uncover the information acquired by the model in pre-training even after additional training seem to have covered it. }
    \label{fig:math_diff}
\end{figure}

\begin{figure}[h!]
    \centering
    \includegraphics[width=0.9\linewidth]{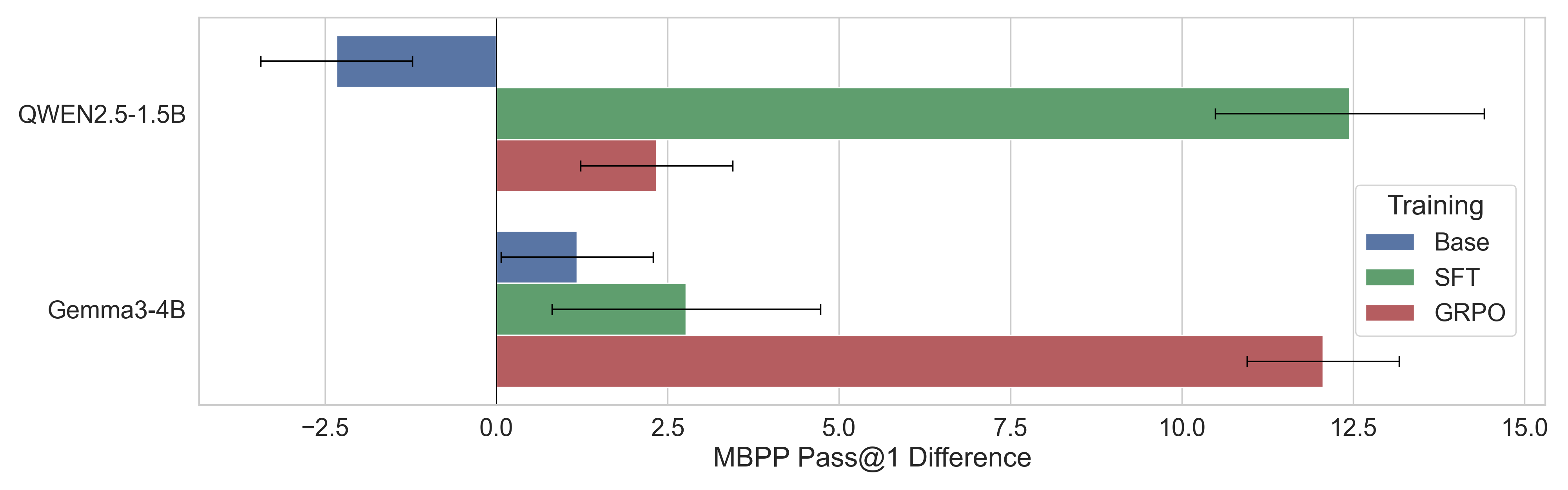}
    \caption{Performance Difference on Code: Accuracy difference between Contaminated and Clean models right after pre-training(base) and after the SFT and GRPO steps on the MBPP benchmark. The same trend with Figure \ref{fig:math_diff} holds for the code benchmarks as well. Here we only present the larger models as the smaller non-instruction tuned models had noisy evals for the coding benchmarks.}
    \label{fig:code_diff}
\end{figure}

Initially, we examined how the contaminated and clean model performance changes over the training process. Specifically in Figure \ref{fig:contam_temp}, we present Qwen2.5-1.5B's contaminated and clean accuracies on the GSM8K benchmark. Similar to previous work \citep{kocyigit2025overestimationllmevaluationcontrolled}, we observe that performance of the contaminated model spikes at the time of contamination then decreases back to the same level as the clean model. This observation is also supported by the base results shown in Figure \ref{fig:math_diff} and \ref{fig:code_diff}. For the model families and benchmarks that we consider in our experiments, five copies of the test set in a pure pre-training setting does not seem to cause measurable and consistent performance inflation compared to an uncontaminated base model. 

In Figure \ref{fig:math_diff} and \ref{fig:code_diff} we also present the performance difference between the contaminated and clean models after the specific post-training step. Here we show that while the baseline pre-trained model shows smaller improvements from contamination, post-training can increase the measured performance gap. After post-training the performance gap is above 2\% for all models for both the math and coding benchmarks, except for Qwen2.5-0.5B SFT on GSM8k. The performance gap reaches 4\% for the smallest Qwen2.5-0.5B model. Error bars show 95\% confidence intervals for the (Contaminated--Clean) difference, obtained by combining the per‑dataset standard errors of the contaminated and clean scores via standard error propagation (assuming the two estimates are independent). These intervals quantify evaluation‑time uncertainty from finite test sets (and any decoding randomness, if sampling is used) and do not include training‑seed variability; the exact formula is given in the Appendix \ref{appx:train_eval_specs}.

This suggests that while continued pre-training after contamination masks the advantage acquired by the contaminated model, the information is not forgotten and can be uncovered with task-specific fine-tuning of the model. We also observe that with the exception of Qwen2.5-0.5B, SFT always causes a larger performance gap for the contaminated model compared to GRPO. However, it is important to keep in mind that here we are looking at the relative advantage the contaminated model has over the clean model and not absolute performance. Notwithstanding, we can draw the conclusion that SFT uncovers that impact of contamination in the pre-training comparatively better compared to GRPO for most models we experiment with. For the small Qwen2.5 model, GRPO does not just increase the gap more, it improves the absolute score more than SFT, as well. While similar results have been shown for vision language models \citep{chen2025sftrlearlyinvestigation},this seems to be an exception in our case and SFT seems to work better for larger models in general.

\subsection{Are performance over-estimations actual over-estimations?}
In this context, it is reasonable to question whether injecting the high-quality test set into the pre-traning mixture prompted with the same evaluation prompt could just help the model become a better model and the gap might not be an over-estimation. 

To check if the performance improvements after post-training are actually over-estimations or if they translate to improvements on uncontaminated benchmarks as well, we compare the performance of the contaminated and clean model's performance on benchmarks that aim to measure the same underlying ability. For math, we chose the GSMPlus benchmark and for coding we used HumanEval. 

Results revealed another interesting pattern between SFT and GRPO. When comparing the base (just pre-trained) markers with the SFT markers, we observe that the average movement is horizontal. This means that while SFT introduces larger performance gaps due to contamination, the impact of contamination on an external benchmark remains constant. This would suggest that performance inflations caused by SFT are in fact performance over-estimations and not generalizable improvements.

On the other hand, when comparing the Baseline markers with the GRPO markers we observe a diagonal movement, meaning that the performance gap between the contaminated and clean model grow for both the contaminated and the uncontaminated models. This suggests that GRPO can extract generalizable improvements from the contamination that is included in the pre-training. We suspect that this is a combination of higher quality data and that the evaluation prompt is used exactly in the contamination inserted in the pre-traning. 

\begin{figure}[h!]
  \centering

  \begin{subfigure}[b]{0.49\textwidth}
    \includegraphics[width=\linewidth]{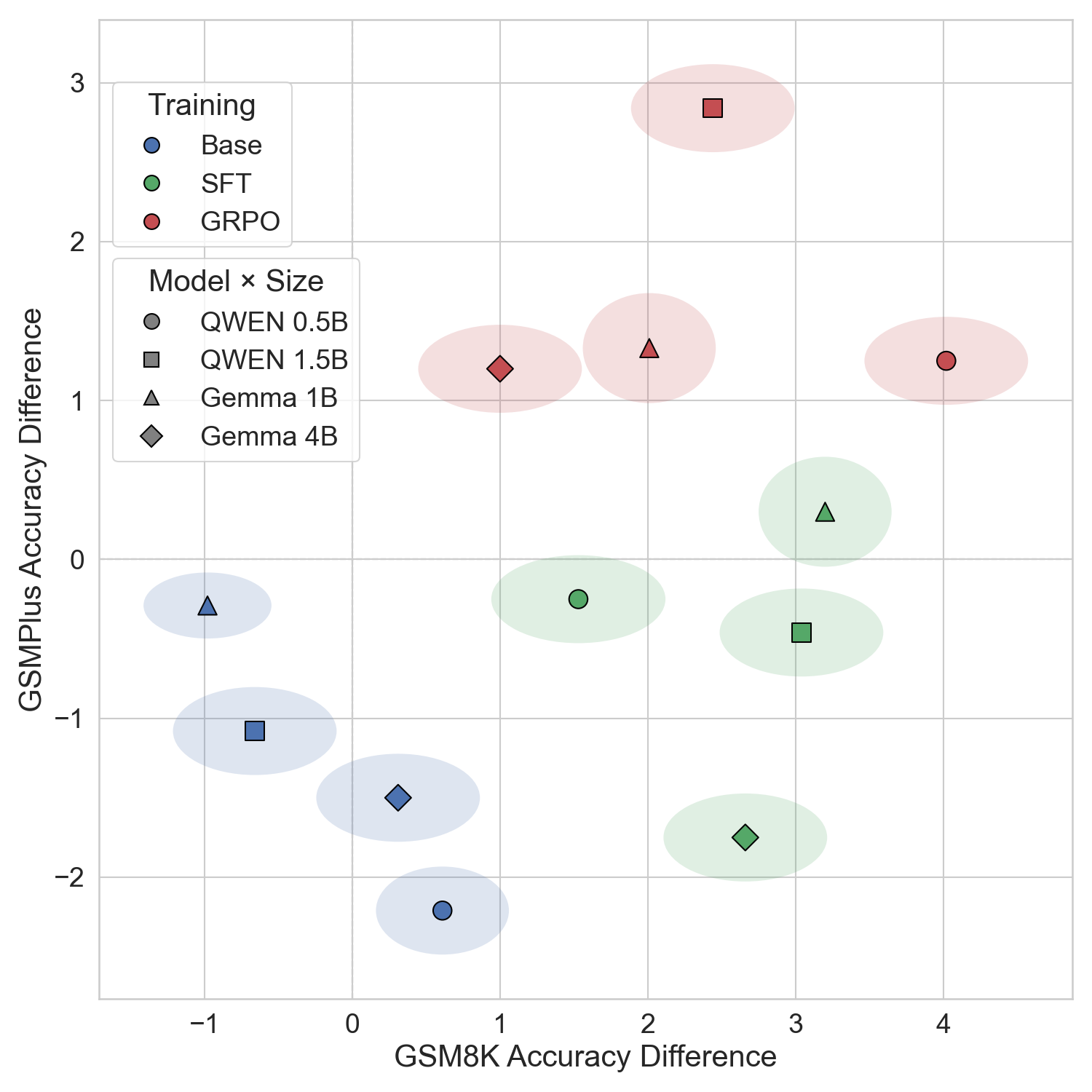}
    \caption{Math Benchmarks}
    \label{fig:math_contam_scatter}
  \end{subfigure}
  \hfill                 
  \begin{subfigure}[b]{0.49\textwidth}
    \includegraphics[width=\linewidth]{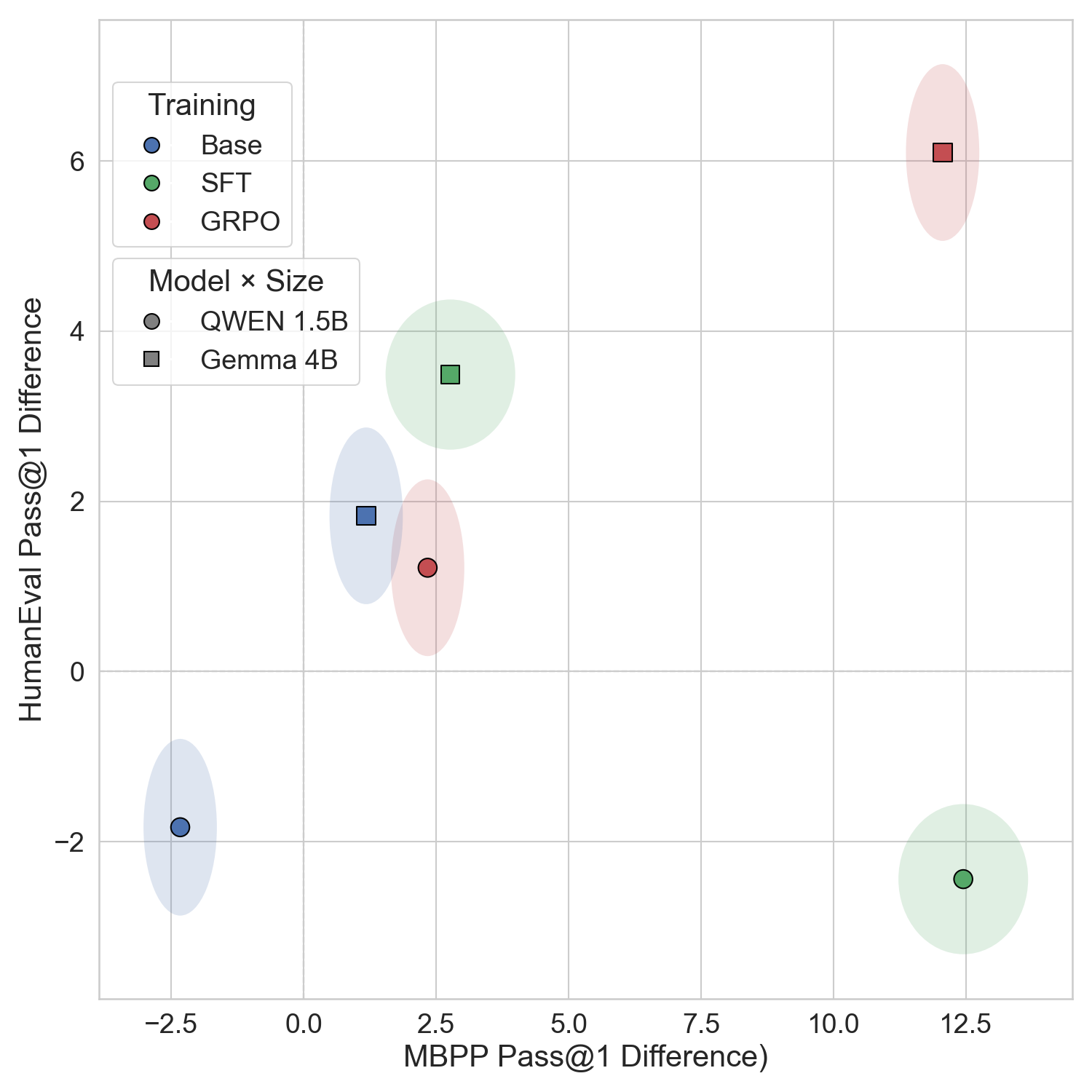}
    \caption{Coding Benchmarks}
    \label{fig:code_contam_scatter}
  \end{subfigure}

  \caption{Comparison of Performance gap on contaminated and uncontaminated datasets. We observe that the Base models behave roughly the same on the contaminated and uncontaminated datasets for both Math and Coding. GRPO fine-tuned models have a positive gap on the contaminated dataset but also have a smaller but still positive gap on the uncontaminated dataset meaning suggesting the models learn some generalizable patterns. The SFT models on the other hand only have a larger gap in the contaminated dataset and show almost no improvement on the contaminated data.}
  \label{fig:contam_scatter}
\end{figure}

For Figure \ref{fig:code_contam_scatter} points mark the performance gap for each model/recipe. Shaded ellipses depict joint 95\% confidence regions obtained similarly to before.

\subsection{How does model scale impact the generalization gap?}
\begin{figure}[h!]
    \centering
    \includegraphics[width=0.9\linewidth]{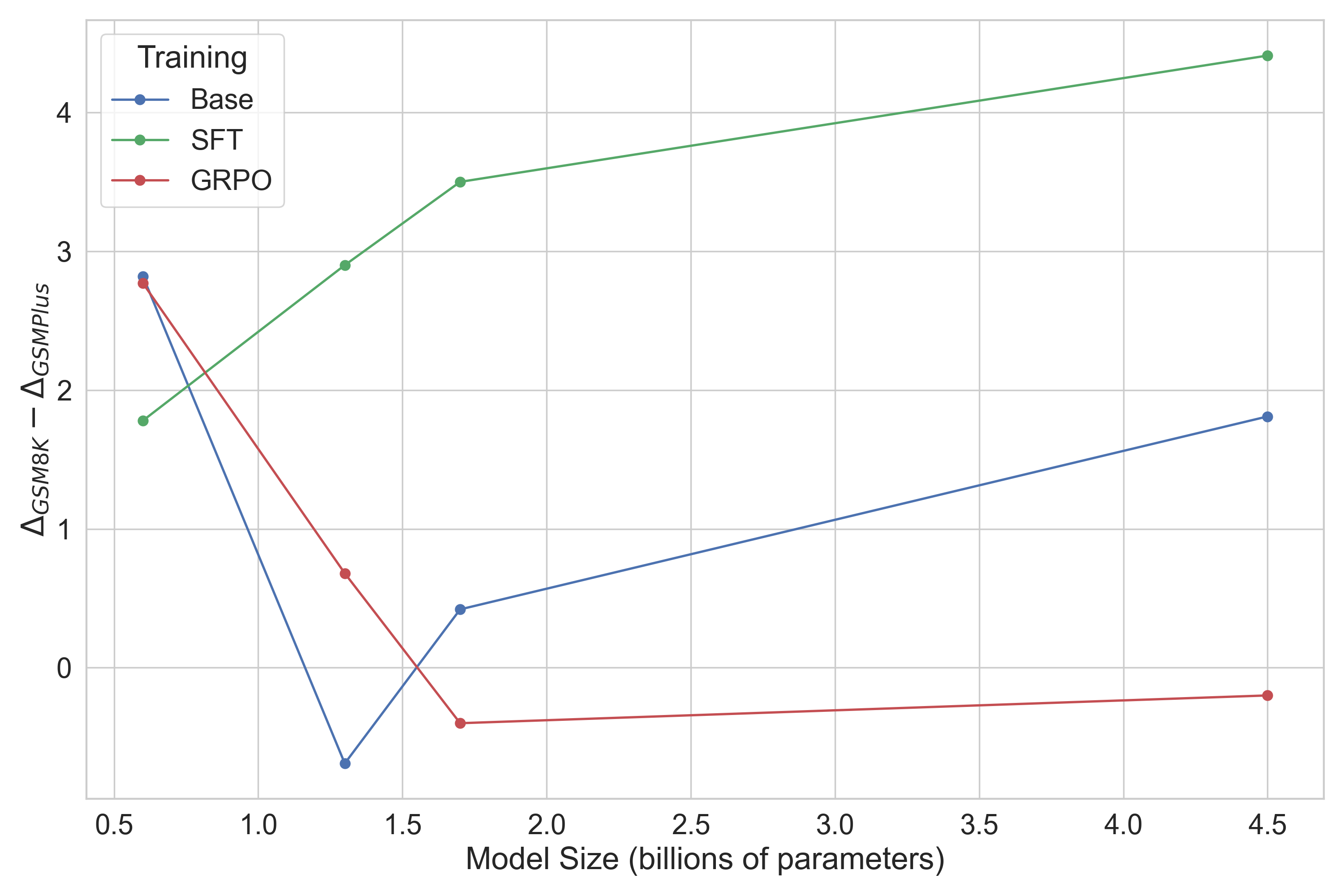}
    \caption{Contamination-gap Difference for Math: We measure the contamination-gap difference as in Eq \ref{eq:delta-gap}. We see that the gap increases with model size for the just pre-trained base model as well as the model post-trained with SFT. However for the models post-trained with GRPO the gap actually gets smaller, even negative as the model size increases.}
    \label{fig:delta_gap}
\end{figure}

To analyze the impact of model scale on model generalization, we now track the \textbf{contamination-gap difference}, defined for each training recipe \(t \in \{\text{Base},\text{SFT},\text{GRPO}\}\) and each model–size pair \((m,s)\) as

\begin{align}
  d_{m,s,t} &\;=\;
  \underbrace{\Delta M_{1}^{\,t}(m,s)}_{\text{GSM8K gain}}
  \;-\;
  \underbrace{\Delta M_{2}^{\,t}(m,s)}_{\text{GSMPlus gain}},
  \label{eq:delta-gap}
\end{align}
where the per-metric contamination improvement is
\begin{equation}
    \Delta M_{k}^{\,t}(m,s)
  \;=\;
  M_{k}^{\mathrm{contam},t}(m,s)
  \;-\;
  M_{k}^{\mathrm{clean},t}(m,s).
\end{equation}
  
where $k\in\{1\text{ (GSM8K)},\,2\text{ (GSMPlus)}\}$. A positive \(d_{m,s,t}\) therefore means the contaminated model gains more on GSM8K than on GSMPlus after recipe \(t\); a negative value indicates the opposite. We present the results in Figure \ref{fig:delta_gap}. Here we see that for the pure pre-trained Base model and the supervised finetuned model, the contamination-gap difference increases as model scale increases, which suggests more over-estimation from contamination. However, for the model post-trained with GRPO, as the model scale increases, the model is actually able to learn more generalizable features and the contamination-gap difference is smaller. 

\section{Discussion}\label{sec:discussion}

Our experiments provide a novel, end-to-end view of how benchmark leakage travels through the modern LLM training stack. We experiment with two model families on two task types and 4 benchmarks. Overall we observe that our findings seems to be mostly consistent across the two model families (Qwen2.5 and Gemma3). Through our experiments three themes emerge:

\paragraph{When you measure matters as much as what you measure.}
Figures \ref{fig:math_contam_temp}–\ref{fig:code_diff} show that the apparent gap between contaminated and clean models almost vanishes once pre-training resumes on fresh data, a result consistent with \citet{kocyigit2025overestimationllmevaluationcontrolled}. Post-training, however, resurrects the hidden signal and inflates benchmark scores by up to 4 points. This finding cautions against relying solely on pre-training checkpoints for contamination analysis and points to the need for \emph{life-cycle} evaluations.

\paragraph{Post-training recipe determines whether leakage over-fits or generalizes.}
Both SFT and GRPO widen the gap on the contaminated task, yet they diverge sharply on uncontaminated benchmarks (Figure \ref{fig:contam_scatter}). SFT’s gains are almost purely local: the contaminated model answers GSM8K or MBPP questions better compared to the base model, but exhibits no extra competence on GSMPlus or HumanEval.  GRPO, in contrast, creates a gap between these two models on both the contaminated and the uncontaminated dataset. This potentially suggests that it learns more generally useful reasoning patterns, yielding a smaller but detectable boost on the clean tasks. 

\paragraph{Scale amplifies the contrast.}
Section \ref{sec:results} and Figure \ref{fig:delta_gap} reveal that larger SFT models extract \emph{more} benefit from contamination that does not translate into a relative benefit on non-contaminated benchmarks. GRPO shows the opposite trend: bigger models channel additional capacity toward broad generalization and thus \emph{dilute} relative over-estimation. The interplay between scale and alignment methods, therefore, deserves careful attention, but our findings are in line with \citet{chu2025sftmemorizesrlgeneralizes}

\subsection{Limitations and Future Work}

Our study purposefully isolates a \emph{5-copy, late-injection} contamination scenario.  Real-world leakage can be multi-pass, paraphrased, or distributed throughout pre-training, and its effects may differ. Moreover, we restrict model sizes to 4B parameters and focus on two open-weights families. That said, larger proprietary models, different architectures, or alternative RLHF algorithms could behave differently. Finally, our RL setting uses synthetic rule-based rewards; human-annotated preference signals might have different effects.

With these limitations in mind, an immediate first step for future research is to run our experiment on a larger scale. While we share some insights into how our results would scale, the analysis is still very local and there are potentially mixed signals that would benefit more from further exploration. Additionally, with more and more complex RL systems, the post-training stage can become just as complex as the pre-training stage, warranting further investigations regarding the compute that goes into the post-training stage and the impact it has on model behavior and performance.

\bibliography{custom}
\bibliographystyle{icml2025}


\appendix



\section*{Appendix A — GRPO Details}
\label{appx:rewards}

\subsection*{A.1 Dataset}
Our experiments fine-tune on the \textit{GSM8K} (\verb|openai/gsm8k|, \texttt{main} split). Each prompt begins with a \emph{system} message that prescribes the \texttt{<reasoning>} / \texttt{<answer>} XML schema, followed by a single worked example (\emph{``What is 2 + 2?''} → 4), and finally the user question drawn from GSM8K.  The gold integer answer is extracted from the dataset for the correctness reward.

\subsection*{A.2 Model and Tokenizer}
The script supports any causal-LM checkpoint that fits on GPU, and has been validated on Google’s \textit{Gemma-3-1B-IT} and \textit{Gemma-3-4B-IT} as well \textit{Qwen2.5-0.5B-IT} and \textit{Qwen2.5-1.5B-IT}. The tokenizer comes from the same directory; we set the padding token equal to
\verb|\eos| and enable a beginning-of-sequence token. This particularly impacts the performance of Gemma3 models.

\subsection*{A.3 GRPO Training Configuration}
We train with GRPO (Group Regularised Policy Optimisation) for a single epoch over GSM8K train set.  Optimisation uses 8-bit AdamW. The model produces sixteen candidate generations for each input example.

\subsection*{A.4 Reward Functions}
Policy updates rely on lightweight heuristic rewards, each returning a scalar per generated sample; their contributions are summed without additional weighting. The complete Python
implementation is reproduced below.

\begin{lstlisting}[language=Python, caption={Reward functions used by GRPO}]

# -- Math Reward functions --------------------------------------
def format_reward_func(completions, **_):
    """1 pt for a perfectly formatted XML response."""
    pattern = r"^<reasoning>(?:(?!</reasoning>).)*</reasoning>\n" \
              r"<answer>(?:(?!</answer>).)*</answer>$"
    responses = [c[0]["content"] for c in completions]
    return [1.0 if re.match(pattern, r) else 0.0 for r in responses]

def correctness_reward_func(prompts, completions, answer, **_):
    """2 pt when <answer> matches the gold integer."""
    responses = [c[0]["content"] for c in completions]
    preds     = [extract_xml_answer(r) for r in responses]
    return [2.0 if p == a else 0.0 for p, a in zip(preds, answer)]

def int_reward_func(completions, **_):
    """0.5 pt if <answer> contains any integer."""
    responses = [c[0]["content"] for c in completions]
    preds     = [extract_xml_answer(r) for r in responses]
    return [0.5 if p.isdigit() else 0.0 for p in preds]

def strict_format_reward_func(completions, **_):
    """0.5 pt for newline-strict XML formatting."""
    pat = r"^<reasoning>\n.*?\n</reasoning>\n<answer>\n.*?\n</answer>\n$"
    responses = [c[0]["content"] for c in completions]
    return [0.5 if re.match(pat, r) else 0.0 for r in responses]

def soft_format_reward_func(completions, **_):
    """0.5 pt for a laxer XML pattern (tags may abut)."""
    pat = r"<reasoning>.*?</reasoning>\s*<answer>.*?</answer>"
    responses = [c[0]["content"] for c in completions]
    return [0.5 if re.match(pat, r) else 0.0 for r in responses]

def xmlcount_reward_func(completions, **_):
    """Fractional reward based on correct tag usage."""
    responses = [c[0]["content"] for c in completions]
    return [count_xml(r) for r in responses]
\end{lstlisting}


For the code specific post-training the main difference is the reward models that are used 

\begin{lstlisting}[language=Python, caption={Reward functions used by GRPO}]

# -- Code Reward functions --------------------------------------
def tests_reward_func(completions, *, tests: List[str], **_) -> List[float]:
    rewards = []
    for c, t in zip(completions,tests):
        code = clean_completion(c[0]["content"])
        #print('--' * 20)
        #print('Running code:', code)
        try:
            passed, _, _ = run_with_timeout(code, t, time_limit=1.0)
            rewards.append(2.0 * passed / len(t))        
        except Exception:
            rewards.append(0.0)
    return rewards


def typehint_reward_func(completions, **_) -> List[float]:
    codes = [c[0]["content"] for c in completions]
    scores = []
    for code in codes:
        try:
            tree = ast.parse(code)
            hints = sum(bool(a.annotation) for a in ast.walk(tree)
                        if isinstance(a, (ast.arg, ast.FunctionDef)))
            scores.append(min(0.25, 0.05 * hints))  # cap at 0.25
        except Exception:
            scores.append(0.0)
    return scores

def brevity_reward_func(completions, **_) -> List[float]:
    codes = [c[0]["content"] for c in completions]
    return [max(0.0, 0.25 - 0.002 * len(code.splitlines())) for code in codes]
\end{lstlisting}


\section*{Appendix B — Prompt Templates}
\label{appx:prompt_templates}

\subsection*{B.1 Training Prompt - GRPO}

Each training sample is a sequence of four chat messages.  
The placeholder \texttt{\{QUESTION\}} is substituted with the GSM8k problem text; the corresponding ground-truth integer answer is kept separately for reward computation.

\begin{verbatim}
<system>
A conversation between User and Assistant. The user asks a question,
and the Assistant solves it. The assistant first thinks about the
reasoning process in the mind and then provides the user with the
answer. The reasoning process and answer are enclosed within
<reasoning> </reasoning> and <answer> </answer> tags, respectively,
i.e., <reasoning> reasoning process here </reasoning>
<answer> answer here </answer>. The answer must be a single integer.
Format example:
<reasoning>
...
</reasoning>
<answer>
...
</answer>
</system>

<user>
What is 2+2?
</user>

<assistant>
<reasoning>
To calculate 2+2, we simply add the numbers together: 2 + 2 = 4.
</reasoning>
<answer>
4
</answer>
</assistant>

<user>
{QUESTION}
</user>
\end{verbatim}

\subsection*{B.2 Training Prompt - SFT}
For SFT we take the training split of GSM8k and process the question, chain of thought and answer separetely. The model is trained on the prompt structure below but all the tokens before let's think step-by-step are masked to not produce any loss. Thus the model is only trained on the chain of thought and answer given the input question. 

\begin{verbatim}
Question: {QUESTION}\
Let's think step by step.
{CHAIN_OF_THOUGHT}.
The answer is {ANSWER}
\end{verbatim}

\subsection*{B.3 Evaluation Prompt}
All evaluation follows the same format where we just prompt the model with the question followed by a "Let's think step by step." primer. We opted for zero shot evaluations as the few shot examples can impact how we have contaminated the data as well. 

\begin{verbatim}
Question: {QUESTION}\
Let's think step by step.
\end{verbatim}

\section*{Appendix C - Training and Evaluation Specifics}
\label{appx:train_eval_specs}

Error bars in Figure \ref{fig:contam_temp}, \ref{fig:math_diff} and \ref{fig:code_diff} denote 95\% confidence intervals for the (Contaminated--Clean) difference, computed by propagating per‑dataset standard errors of accuracy estimated from \(n\) test items \(\bigl(\mathrm{SE}=\sqrt{\hat{p}(1-\hat{p})/n}\bigr)\); the half‑width is \(1.96\,\sqrt{\mathrm{SE}_{\text{contam}}^{2}+\mathrm{SE}_{\text{clean}}^{2}}\) (independence assumed). These intervals reflect evaluation‑time uncertainty only and exclude training‑seed variability.

\end{document}